# A Perspective on Sentiment Analysis

K Paramesha[a,] a*, K C Ravishankar[b]

[a]Department of CSE, Faculty of Technical education, Vidyavardhaka College of Engg, Mysore,570023,India
[b]Department of CSE, Faculty of Technical education,Government Engg College, Hassan,573201,India

**Abstract**

Sentiment Analysis (SA) is indeed a fascinating area of research which has stolen the attention of researchers as it has many facets and more importantly it promises economic stakes in the corporate and governance sector. SA has been stemmed out of text analytics and established itself as a separate identity and a domain of research. The wide ranging results of SA have proved to influence the way some critical decisions are taken. Hence, it has become relevant in thorough understanding of the different dimensions of the input, output and the processes and approaches of SA.

## 1. Introduction

The Web is getting fattening day by day with tons of data posted on various matters happening every now and then. With a phenomenal and exponential growth of data, the manual processing of textual data so as to extract the knowledge is totally ruled out simply because of feasibility and cost effective issues in capturing and comprehending such a humongous amount of data. Interestingly, with the advancement in the computational speed of processors, which are getting doubled every year, automated processing of data would be accomplished by design and development of sophisticated, intelligent programs. Eventually, leads to the development of innovative tools and techniques to derive the hidden knowledge for decision making in corporate and government organizations.

Such processing tasks are collectively referred as Text Mining or Text Analysis, involving the following tasks:

- Characterization Task
- Annotation Task
- Extraction Task
- Summarization Task
- Classification Task

## 2. What is Sentiment Analysis?

Sentiment analysis [1, 6, 8] is a specialized branch of text mining, which involves one or more combination of the above mentioned tasks. It is an automated process to extract opinion bearing phrases in a piece of text or classifying a piece of text into positive or negative classes by the application of one or more combination of statistical, linguistic, machine learning and natural language processing tools and techniques.

* K Paramesha. Tel.:+919449059697; fax: +91 821 2510677.
  *E-mail address*: paramesha.k@vvce.ac.in, kcrshankar@gmail.com.

K Paramesha .et.al.

## 3. Types of SA

Basically SA is a classification problem which classifies a given review document to positive or negative polarity. Over the recent years, it has grown beyond our imagination and the needs of the practical applications are growing steadily. In-depth, more detailed and diversified analyses [6] suitable for generating the desired outcome for a given problem is absolutely essential. Here we captured some of the important types and levels of SA.

*3.1 Coarse Grain SA*

In this type of analysis, given a review document or review paragraph on an object say, a DVD or a cell phone, it is required to determine the inclination of input text whether positive or negative or neutral and also some authors have considered the subjectivity analysis at sentence level as coarse grain analysis [6, 17].

Consider the following review text on iPhone mobile. Determining, whether it is positive or negative attitude of the author on iPhone, or computing positive or negative feeling of each sentence, is said to be coarse grain sentiment analysis.

*(1) I bought an iPhone a few days ago. (2) It was such a nice phone. (3) The touch screen was really cool. (4) The voice quality was clear too. (5) Although the battery life was not long, that is ok for me. (6) However, my mother was mad with me as I did not tell her before I bought it. (7) She also thought the phone was too expensive, and wanted me to return it to the shop.*

*3.2 Fine Grain SA*

In Contrast to the coarse grain, in-depth analysis at feature/aspect level can reveal more and precise information [6, 37, 50, 55, 59, 70] since sentiments are expressed on a single topic or multiple topics in a single sentence or across sentences. Extracting opinions at phase level or sentence level on topics and subtopics are also referred as feature or aspect oriented sentiment analysis. Here, if topic is denoted as an object, then the subtopics are the components of the object and having part–of relation with the object. In the above example, a topic is the iPhone mobile, which has the parts such as "touch", "screen" and "battery" referred as subtopics on which sentiments are expressed. The figure 3.1 shows the plot of output summaries of features of two cell phones.

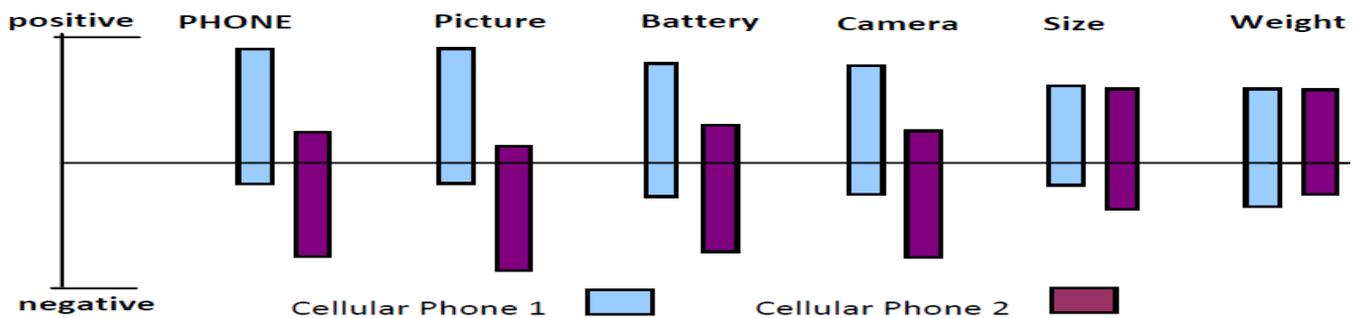

Fig. 3.1 Feature-based opinion summaries of two cell phones

*3.3 SA Using Emotional Connotation*

Recently, an emerging task of detection of emotional connotation in text is gaining importance and promises the economic stakes [30, 58, 68]. For example, Postings thrown at blogosphere, forums and social networking sites, contains great deal of information. Thus, computation on such data can shows trends in user moods across the globe. Consequently, affecting other correlated events, like stock market prices, electoral polls. It is a task of annotating a piece of text using a predefined list of emotions: anger, disgust, fear, joy, sadness and surprise. Further, the valence, positive or negative, for each of the emotions is determined to say the attitude/feeling of the text. The Table-I gives the polarity association for the emotions.

Table-I. Polarity amplification for emotions

| Words | Emotions | Polarity |
| --- | --- | --- |
| Aggression Hostility Wrongful Conduct | Anger, Fear, Sadness, Disgust | Negative |
| Weaponry, Weapon System | Anger, Fear, Sadness | Negative |
| Welcome,Perks,Incentives,Prize | Joy | Positive |

*3.4 Domain Specific and Cross Domain SA*

A domain in sentiment analysis could be informally characterized by the set of features, lexicons, style and syntactic way of expressing the opinions, each of which varies across domains. Some of the deemed domains are Products, Books, Movies, Political Discourse Newsgroups, Automobiles, Banks, Travel Destinations. Intuitively, sentiments are often expressed with domain-specific vocabulary, and also words which give different sentiments across several domains. For example,

The market rises in the afternoon with inflow of huge investment

The heat beat rises above the normal rate when this medicine is administered

The word rises indicate the positive attitude in the stock market domain, whereas the same word gives a negative feeling in the medicine domain. Determination of domain-specific clues and their polarity is more challenging one as compare to domain independent sentiments. There are some lexicons unique to a specific domain and others common across all the domains. Building domain-dependent lexicons for many domains is much harder work than preparing domain independent lexicons and syntactic patterns, because the possible lexical entries are too numerous, and they may differ in each domain[64, 69].

## 4. SA in Opinion Search and Retrieval

Opinion retrieval and search is a subsidiary task of search engines in rating opinions and trends, and enriching the search experience of users. Like a search engine, which, given a topic or a phrase, retrieves the web pages and documents on web which are relevant to the input query and orders them based on the relevance, the opinion search engine retrieves the opinions expressed on the input phrase after performing the sentiment analysis, and then presents the results in the order of significance. Note that the positive, negative sentiments are retrieved and their proportions are computed [1, 4, 7 ].

*4.1 SA in Opinion Spam and Utility*

Sometimes reviews are written to promote the product which is not good as opposed to what it claims to be in reviews, and also reviews are written to damage the reputation or to degrade a product or services which are actually better than what it is said about them in reviews [6, 7, 8]. Both cases are disseminating false information and are spam. To get rid of such spam reviews, which have no utility, we may perform sentiment analysis to detect spam reviews and isolate them from other legitimate reviews. Practically it is highly impossible to remove spam and requires a complex process as the reviews can be written in so many ways and styles [62, 64, 69].

*4.2 SA in Sentiment Lexicon Building*

Identifying sentiments of words in review input requires a huge sentiment lexicon dictionary which gives senses along with the words [7, 62]. It is desired to capture sentiments of opinion bearing words, failing which can affect the performance. In order to build such a large sentiment lexicon, several approaches have been proposed, namely manual approach, dictionary-based approach, and corpus-based approach. For instance, a product lexicon dictionary can be developed to aid capturing sentiments of all electronic products [6, 16, 25, 47, 71].

## 5. Input Text Data Characteristic Features

The design of processing model of the sentiment analysis algorithm is determined by the nature of the input opinionated text data and the required output that has to be generated. Naturally, it is imperative to understand the different ways and types in which input opinionated text document contents can be represented, and to learn their characterization as well.

Any input text data contains objective statements (about facts) and subjective statements (opinion expressed about something). A statement written using natural language, say English, can be ambiguous as the nature of language itself is like so. It is also worth to mention that there are differences of opinion among different people about a statement whether it is a subjective or objective [1, 6, 7].

Sentence1: *The sunset is beautiful and spectacular.*
Sentence2: *After sunset, birds flew back to their nests.*

In the sentence1, there is an opinion or private state about the sunset and the opinion is positive. Such opinion bearing words and phrases can be expressed literally on anything. For example, on products, services, objects and so on. In contrast with the sentence1, sentence2 reports the factual information and does not bear private state.

We now introduce some of the technical words used in and around the sentiment analysis as it is helpful in understanding the subject in detail.

- **Object:** An *object o* is an entity which can be a product, person, event, organization, or topic.
- **Opinion Phrase:** An *opinion phrase* on a feature of an object in a sentence that expresses a positive or negative opinion
- **Opinion:** An *opinion* on a feature is a positive or negative view, attitude, emotion or appraisal on *the feature.*
- **Opinion Holder:** The *holder* of an opinion is the person or organization that expresses the opinion.

## 5.1 Features and Attributes In Clauses

A review text on an object may contain opinionated text on the object itself and on its components and subcomponents since an object is made up of several components and subcomponents [1, 6, 48, 50, 55, 59, 70, 76]. Consider the following review on a computer (Electronic Products)

*This computer is amazing!! The 4 GB of RAM makes it fly when using any program and the hard drive is huge. There is enough room to store my whole music collection plus many photos. The Wi-Fi connected to the internet instantly and the Bluetooth helped me transfer my pictures and videos from my phone onto my computer.*

In the above review, the computer is a whole part and RAM, Screen, Battery, Wi-Fi, Bluetooth, CD/DVD Drive and Color are some of the components / features. Likewise, the design, color, size, battery life are some of the attributes of the computer and its parts.

## 5.2 Explicit and Implicit Clauses

The sentences or phrases, expressing opinions on objects or features which are stated directly in clause are said to be explicit [1, 6, 42, 62, 64, 71]. However the features, which are comment on, are recognized without any ambiguity. In the following example:

*My one complaint is that the slot load CD/DVD drive is very noisy.*

The CD/DVD is the explicit feature in the sentence. Where as in the following sentence,

*There is enough room to store my whole music collection plus many photos*

The feature is not directly stated, but rather implicit, which is actually, "Hard disk space" feature.

## 5.3 Positive, Negative, Mixed and Neutral Clauses

The sentences present in the input document reflect different emotions. While performing analysis at the sentence level, it is important to detect which class each sentence belongs to. The following sentences show the example for each category type.

POSITIVE:   *Ishmael presents a remarkably different view of the history of mankind.*
NEGATIVE:   *The story didn't seem to be very original.*
MIXED:      *The book links all questions rather beautifully but still fails to impress.*
NEUTRAL:    *What is true for one man is not always true for rest of us.*

In all of the cases [15, 45, 52, 62], the different opinions are expressed on varied number of features in the same sentence. So, it is important to know which appraisal is on which feature. Looking at the below statement, opinions could be expressed on two features "voice quality" and "battery life" in same sentence which is called as compositional statement.

*The voice quality of this phone is good, but the battery life is short*

## 5.4 Comparative Clauses

Sometimes opinions are expressed in terms of comparisons. Objects or features are compared with other objects or features which are similar. The comparisons are show to bring out the distinctions between former object / feature and later object/feature, and to say which one is better/bad than the other with respect to something [6, 8, 33, 64, 71]. A comparative opinion is usually expressed using the *comparative* or *superlative* form of an adjective or adverb, although not always. The compared objects/features would of the same brand or different ones.

## 5.5 Negation Clauses

Sentences in the input can be thought of two types, affirmative and negative (non-affirmative) sentences [5, 18, 32, 35, 45]. It is a very common linguistic construction that affects polarity and, therefore, needs to be taken into consideration in sentiment analysis. In the following statement the negation is straight forward. This need not be the case always.

*I do not like this new Nokia model*

In some case, although there are negative terms in the sentence, they do not alter the polarity of the statement. In the following sentence, the negation word "not" used in construct does not make the statement negative. However, it is not always that easy to spot positive and negative opinions in text having negations.

*<u>Not</u> only is this phone expensive but it is also heavy and difficult to use.*

*5.6 Conditional and Interrogative Clauses*

Conditional statements are use to show cause-effect relations or a hypothetical situations [1, 4, 6, 68]. For example, in the below statement, although there are sentiment words "higher", "break" and "opponent", the sentence is completely neutral(no sentiments).

*If your card's number is <u>higher</u> you can <u>break</u> the <u>opponent</u> card*

Interrogative statements are very challenging to evaluate the polarity. Considering the below sentence, It is neutral even though word "wrong" is present.

*What's wrong with the pilot?*

If the same sentence is rewritten like below, the sentence polarity changes from neutral to negative about the pilot

*What's wrong with the i<u>nexperience</u> pilot?*

*5.7 Thwarting Clauses*

In the beginning of review, the author of the review creates positive impression about the product/features and it is believed that author is fully convinced and positive, but interestingly, the author in the end takes u-turn on the closing note and eventually the review turn out to be negative[1, 6, 62]. In the following review, the last sentence makes the entire review negative.

*This film should be brilliant. It sounds like a great plot, the actors are first grade, and the supporting cast is good as well, and Stallone is attempting to deliver a good performance. <u>However, it can't hold up.</u>*

# 6  Challenges in SA

Literature on SA and its evolution clearly shows that most of the sentiment tasks and sub tasks are still open-ended problems [1, 4, 6, 7]. The main reason for having abundant scope and challenges is that we are not dealing with numbers! Instead it is the textual information we are working with, which has several properties and characteristics within it. Based on this, we present few important challenges in SA.

*6.1 Recognizing Subjective Expression*

Subjective expressions come in many forms, e.g., opinions, allegations, desires, beliefs, suspicions, and speculations [1, 2, 6, 9, 11, 64, 71]. Thus, a subjective sentence may not contain an opinion. For example,

*I want to buy a good cell phone.*

The sentence is subjective but it does not express a positive or negative opinion on any specific phone. Similarly, we should also note that not every objective sentence contains any opinion as the second sentence. Detection of right candidate of subjective expression that most annotators may be agreed upon is indeed the most challenging one simply because the thin line separating subjective and objective expression and the task is even more complicated than the subsequent phase of detecting the polarity of the expression.

*6.2 Recognizing the Context Polarity of Opinionated Words*

We know that a given word is positive or negative in its sense which is nothing but prior polarity [2, 4, 5, 11, 45]. However, the prior polarity of a word changes with the context. In other words, the contextual polarity of the opinion phrase in which a particular instance of a word appears may be quite different from the word's prior polarity. Negative words are used in phrases expressing Positive sentiments, or vice versa. Also, quite often words that are positive or negative out of context are neutral in context, meaning they are not even being used to express a sentiment.

In the below sentence, the "weak enemies" shall not be considered as expression of sentiment, but rather a neutral expression because they are talking about a game in which they should begin playing against weaker side (weaker enemies). Finally, all that matters is the context in which the words are used

*There is a steep learning curve in this game, which can only partly be compensated by starting off with <u>weak enemies</u>.*

In the example sentence below, the first underlined word is "trust." Although many senses of the word "trust" express a positive sentiment, in this case, the word is not being used to express a sentiment at all. It is simply part of an expression referring to an organization that has taken on the charge of caring for the environment. Out of context, words "reason" and "reasonable" are

Positive prior polarity, but in context, the word "reason" is being negated, changing its polarity from positive to negative. The phrase "no reason at all to believe" changes the polarity of the proposition that follows; because "reasonable" falls within this proposition, its polarity becomes negative.

*Philip Clapp, president of the National Environment Trust, sums up well the general thrust of the reaction of environmental movements: "There is no reason at all to believe that the polluters are suddenly going to become reasonable."*

### 6.3 Recognizing Negations in the Phrases

Automatic detection of linguistic negation in sentences and phrases is a critical need for many text processing applications, including sentiment analysis. Failing of which will affect the performance of the system [4, 5, 6, 8, 10, 45]. Unfortunately, sentences of negation are not always so syntactically simple. The negation phrase construction can be complex and negative words appearing in phrases may not always need to negate the sentences. At the highest structural level, negations may occur in two forms: 1) morphological negations, where word roots are modified with a negating prefix (e.g., "dis-", "non-", or "un-") or suffix (e.g., "- less"). 2) Syntactic negation, where clauses are negated using explicitly negating words or other syntactic patterns that imply negative semantics. For example, the below sentence has the negative word "never" has impact on the sentiments expressed.

*<u>Never</u> before in my career, had I enjoyed such a wonderful and blessed day.*

The most common form and are typically unambiguous negations of a particular clause, such as the following sentence.

*The audio system on this television is <u>not</u> very good, but the picture is amazing.*

### 6.4 Recognizing Implicit Feature

Sentiment analysis is not confined in knowing positive or negative sentiments but also to find other associated information such as object, feature, opinion holder, opinion and time at which the opinion is expressed. The detection of implicit features is not an easy task as the features are not stated directly with in the phrases [11, 42, 55, 59, 69].
In the following review on book, the feature "story" of the book is commented on. Since features change domain to domain, real world knowledge of domain is absolutely needed to identify implicit features.

*I loved the strong, independent main character.*

### 6.5 Lack of Annotated Dataset

Since sentiment analysis has come to existence, it has seen tremendous growth in both research and practical applications and is still growing. Even though it is subcategory of the text analysis, it spans a huge length and breath. To perform various types of Sentiment analysis with some specific conditions, and to validate the end results, it is required to use wide range of labeled datasets, which not available easily suitable for the application [1, 6].

For instance, we require dataset in a domain labeled with each and every sentence about the comparison to perform fine-grain sentiment analysis on comparative sentences in the particular domain. However, there are quite a few data sets with limited number of example set.

## 7. Approaches to SA

SA is a multi dimension problem which requires different perspective in techniques and tools to get the desired solution for a specific problem. Based on our observations, we broadly stated the approaches under three categories, comprehensively covering all approaches.

### 7.1 Machine Learning Approaches

Machine learning approaches are based on data model which can be classified into generative and discriminative models. The generative class of algorithms requires the distribution of data. Having known all the distribution, for a new data point x, we compute the probability of generating the x for each label y by applying Bayesian rules and then $\arg_{max} P(x|y)$ gives probable label of x. Naïve Bayes [35, 40, 41, 42], latent dirichlet allocation[19], HMM[43] and mixture models are some of the generative algorithms.

On the other hand, the discriminative class of algorithms straight away computes the boundary that separates different classes using a characteristic equation. SVM [35, 41, 43, 45, 46, 74], conditional random fields [17, 36, 37, 38] and Logistic regression are some of the discriminative algorithms

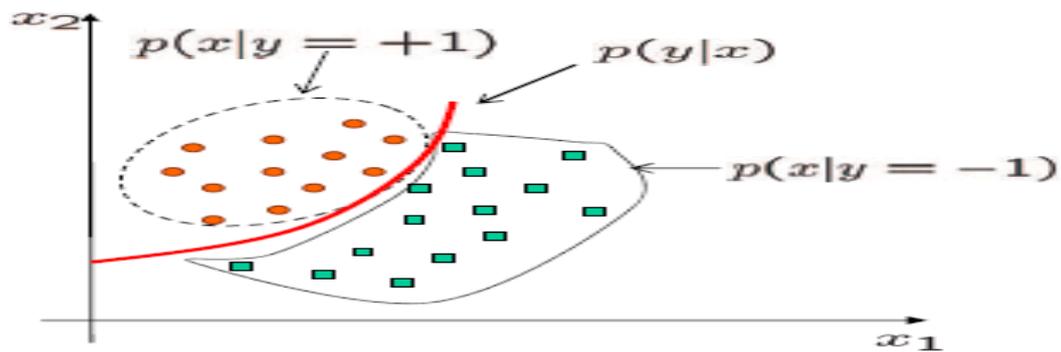

*Fig 6.1 Data distribution and boundary*

*7.2 Statistical and Linguistic Approaches*

It covers a broad range of approaches ranging from term weighting using lexicon resources[10, 16, 18, 25, 52], tfidf weighting[8, 29, 32, 48, 56], computing pointwise mutual information (PMI)[39, 45, 57, 68, 69] using huge corpus, pattern/rule based extraction [23, 63, 65, 66, 68, 75], to computing mutual association of words, dependency relation analysis [34, 59, 65, 67, 74] to extract features and dependency structure. The statistical approaches involve identifying a set of features and converting them into numbers for inference. For instance, frequency of a word (feature), tfidf of a word and length of a sentence are some statistical values. The linguistic ways of feature extraction and inference are based on the language properties. For instance, dependency relations (nsubj, amod, …) between words in a sentence, in rule based approach patterns of the Part of speech (POS) information of words are used to extract word and then decide the polarity in the next course of action in a sequence. In practice, both statistical and linguistic features are determined in decision making.

*7.3 Ontology Based Approaches*

Unlike lexicon resources, ontology is a data structure containing entities and relationships between are modeled for specific application domain. Ontology once built automatically or manually, can be used to perform variety of tasks such as information sharing, information extraction and decision making [23, 51, 53, 56, 57, 63].

## 8. Applications of SA

Applications are numerous and are still growing, to name a few, gathering the public opinions on polls, political and social events to assess the follow-up trends. Evaluating the sentiments expressed on products (such as electronic gadgets, book, movies, software, drugs…), features and services so as to understand the market better and to improve the business propositions. Opinion based search results about something and opinion based spam identification are some of the challenging applications need to be addressed.

## 9. Conclusion

SA for over past few years has grown leaps and bounds. It is observed that a comprehensive tool or methodology is not sufficient for to capture the whole sentiment and for inference of any input textual data. This is mostly due to the inherent linguistic characteristics of data in a natural language. This ultimately makes the NLP task very complex and prone to many issues. Empirically, in order to improve the performance of the system for practical applications, we advocate try and testing of blended set of tasks or approaches to solve a precise and specific problem in a domain.